%% file: main.tex
\documentclass[10pt,twocolumn,letterpaper]{article}
\usepackage[pagenumbers]{cvpr}
\input{preamble}
\usepackage{needspace}
\usepackage{capt-of}
\usepackage{placeins}
\title{CanopyKV: Remember beyond the Crop\\for Fast Active Visual Reasoning}
\hypersetup{pdftitle={CanopyKV: Remember beyond the Crop for Fast Active Visual Reasoning},
pdfauthor={Lei Yang, Dayan Wu, Yiliang Zhu, Xinze Liu, Hengjie Zhu, Zihao Zhang, Ding Wang, Peng Fu, Zheng Lin}}
\author{%
Lei Yang$^{1,2}$\thanks{Lei Yang: \href{mailto:yanglei@iie.ac.cn}{\texttt{yanglei@iie.ac.cn}}.}\quad Dayan Wu$^{1}$\thanks{Dayan Wu: \href{mailto:wudayan@iie.ac.cn}{\texttt{wudayan@iie.ac.cn}}.}\quad Yiliang Zhu$^{2,1}$\quad Xinze Liu$^{1,2}$\\
Hengjie Zhu$^{1,2}$\quad Zihao Zhang$^{1}$\quad Ding Wang$^{3}$\quad Peng Fu$^{1}$\quad Zheng Lin$^{1}$\\[6pt]
{\normalsize $^{1}$Institute of Information Engineering, Chinese Academy of Sciences, Beijing, China}\\
{\normalsize $^{2}$School of Cyber Security, University of Chinese Academy of Sciences, Beijing, China}\\
{\normalsize $^{3}$Department of Applied Mathematics and Statistics,}\\
{\normalsize Johns Hopkins University, Baltimore, MD, USA}}
\begin{document}
\maketitle
\flushbottom
\input{sections/00_abstract}
\input{sections/01_introduction}

\input{sections/02_related_work}
\input{sections/03_formulation}
\input{sections/04_method}
\input{sections/05_experiments}
\input{sections/06_conclusion}
{\small
\AtBeginEnvironment{thebibliography}{\interlinepenalty=10000}
\bibliographystyle{ieeenat_fullname}
\bibliography{references}}
\clearpage
\appendix
\raggedbottom
\setcounter{figure}{0}
\setcounter{table}{0}
\setcounter{equation}{0}
\renewcommand{\thefigure}{S\arabic{figure}}
\renewcommand{\thetable}{S\arabic{table}}
\renewcommand{\theHfigure}{supp.\arabic{figure}}
\renewcommand{\theHtable}{supp.\arabic{table}}
\renewcommand{\theHequation}{supp.\arabic{equation}}
\section*{Supplementary Material}
\input{sections/07_supplement}
\input{sections/09_analysis_archive}
\input{sections/14_selection_gallery}
\input{sections/08_additional_experiments}
\FloatBarrier
\Needspace{29\baselineskip}
\input{sections/16_cache_growth_appendix}
\FloatBarrier
\end{document}

%% file: preamble.tex
\usepackage{algorithm}
\usepackage{algpseudocode}
\usepackage{multirow}
\usepackage{array}
\usepackage{colortbl}
\usepackage{tabularx}
\usepackage{amsfonts}
\usepackage{amsthm}
\usepackage{microtype}
\usepackage{enumitem}
\usepackage{tikz}
\usetikzlibrary{arrows.meta,positioning,calc,patterns}
\definecolor{cvprblue}{rgb}{0.21,0.49,0.74}
\definecolor{avblue}{HTML}{326D9C}
\definecolor{avorange}{HTML}{D1813B}
\definecolor{avgreen}{HTML}{35836A}
\usepackage[pagebackref,breaklinks,colorlinks,allcolors=cvprblue]{hyperref}
\usepackage[capitalize,noabbrev]{cleveref}
\newcommand{\method}{\textnormal{\textsc{CanopyKV}}}

\newcommand{\tablevenue}[1]{{\scriptsize\color{black!65}#1}}

%% file: sections/00_abstract.tex
\begin{abstract}
Active visual reasoning enables multimodal models to answer challenging visual questions by gathering fine-grained evidence through iterative image inspection and tool use. However, accumulating crops and reasoning tokens expands the key--value (KV) cache, making compression important for reducing memory and inference costs. Existing KV compression methods typically select historical tokens based on attention, recency, or redundancy. These criteria do not explicitly distinguish locally replaced evidence from complementary information needed for future exploration: each crop refreshes a local region, while earlier views may still support subsequent decisions. Discarding such evidence can impair accuracy or require additional search, while scoring and cache reorganization can offset the computational savings from compression. We introduce \method{}, a training-free KV compression framework that uses parent--child crop relationships to retain complementary evidence, keeps retained history stable, and protects the state needed for ongoing reasoning. Tool metadata provides retention signals without additional attention scoring, while compact KV storage reduces subsequent attention costs. Across VisualProbe, V*, HR-Bench, and MME-Lite, \method{} achieves state-of-the-art answer accuracy and end-to-end efficiency among the evaluated KV compression methods. On MME-Lite, it retains approximately one-quarter of Dense's total KV while improving accuracy by 10.5 percentage points and accelerating complete reasoning by $1.64\times$.
\end{abstract}

%% file: sections/01_introduction.tex
\input{figures/placeholder_teaser}
\section{Introduction}
\label{sec:introduction}
Active visual reasoning solves difficult visual questions by deciding what to inspect, acquiring new views, and revising its conclusions. Visual programs~\cite{gupta2023visprog,suris2023vipergpt}, guided region inspection~\cite{wu2023vstar,shao2024visualcot}, and learned tool use~\cite{zheng2025deepeyes,wang2025pixelreasoner,lai2025minio3} support this exploration. Earlier observations and visual tool feedback~\cite{hu2024visualsketchpad} provide context for comparing regions, returning to a previous branch, and choosing the next action. Sustaining this capability requires memory that remains useful as the search evolves. Yet every acquired view, tool message, and reasoning step enlarges the key--value (KV) cache, making later steps process an increasingly expensive history (\cref{fig:teaser}a).

Compression changes this exploration as well as its execution cost. Retained history shapes the next decision, which determines the next observation and the work that follows. Keeping everything increases cache reads, while discarding needed evidence can damage answers, disrupt tool use, or prolong search. Thus, a cheaper decoding step need not produce a cheaper completed task. A discarded cue may have to be acquired again through another crop and its associated reasoning. Recovering it adds tool execution, visual encoding, and decoding work, even if each step uses a smaller cache. Cache selection and maintenance add a further cost. The challenge is to reduce memory and execution overhead while preserving the evidence and state needed to complete visual exploration.

Existing KV methods estimate retention value through attention statistics~\cite{zhang2023h2o}, recency~\cite{xiao2023streaming}, text priority~\cite{wan2024lookm}, cross-modal relevance~\cite{tu2025vlcache,huang2025aircache,liu2026tgvkv}, query proxies~\cite{shu2026mmshift}, or semantic diversity~\cite{liu2026mixkv}. These signals estimate relevance or redundancy without explicitly describing how a newly executed visual action changes the role of earlier evidence. In active exploration, history helps determine what the model observes next, so retention must account for both the information supplied by that observation and the context still needed afterward. Evidence unused during an intermediate step may become useful when the model compares regions or returns to another branch. The task is therefore not simply to favor the most recently useful tokens, but to preserve context across changing information needs. Executed tools already record the source and location of each new observation. These records offer a structural cue for selecting useful history without repeatedly extracting model-derived importance statistics.

Our central insight is \emph{local evidence replacement and persistent complementarity}. \emph{Local replacement} means that a new crop can resupply evidence within its footprint, rather than making its entire parent view redundant. \emph{Persistent complementarity} means that unreplaced parent evidence may remain useful across subsequent observations and should stay available for later decisions. In \cref{fig:teaser}a, A1 refines part of A but leaves the parking information in C uncovered. This context can remain useful after intervening observations A2 and B, for example when choosing to inspect C. Panels b and c illustrate whether it survives in parent history at an equal parent-KV budget. The interventions in \cref{sec:evidence-observation} test which parent evidence the new crop can replace, while the delayed-readout studies in \cref{sec:mechanism} examine whether complementary evidence remains usable across branch visits. Together, the two ideas specify what to retain and why retaining it only at the current step is insufficient.

\looseness=1
We introduce \method{}, a training-free framework that makes this complementary history compact and reusable (\cref{fig:framework}). To select useful evidence, source quotas balance the available visual sources and reasoning text, while executed crop--parent relationships, or \emph{crop lineage}, identify context beyond the incoming view within parent crops. Preserving that context at one step is insufficient if it disappears before the search returns. \method{} therefore keeps selected history fixed across later observations. Protected task instructions and recent control state allow the model to continue reasoning from returning tool observations. Together, these choices preserve both the evidence needed by later branches and the state needed to use it.

\looseness=1
The same design keeps this reusable history inexpensive to maintain and use. Selection relies on structure already recorded by visual tools, avoiding compression-specific attention extraction, similarity scoring, and extra model forward passes. Selecting each completed segment once avoids repeatedly ranking earlier history, while compact native KV tensors shorten cache reads throughout subsequent decoding. These savings address both the work of maintaining memory and the repeated cost of using it. Preserving useful evidence therefore supports accurate exploration while reducing the cost of completing the task.

\Needspace{4\baselineskip}
Across five evaluation groups from four benchmarks---VisualProbe, V*, HR-Bench, and MME-RealWorld Lite---\method{} outperforms the compared baselines in both answer accuracy and reasoning speed (\cref{tab:quality}). On MME-Lite, retaining 24.63\% of Dense's total KV improves accuracy by 10.47 percentage points and accelerates complete tasks by $1.64\times$.

\Needspace{6\baselineskip}
Our contributions are threefold.
\begin{itemize}[leftmargin=*,itemsep=2pt,topsep=3pt]
\item We introduce, to our knowledge, the first historical KV compression framework tailored to crop-based active visual reasoning, turning cache reduction into end-to-end acceleration.
\item We propose \method{}, a training-free framework that translates local evidence replacement and persistent complementarity into lineage-aware selection, fixed history, and native KV execution, enabling persistent evidence retention and low-cost reuse.
\item We achieve higher answer accuracy and task efficiency across four benchmarks, with controlled studies supporting complementary evidence retention and low-cost execution.
\end{itemize}
\input{figures/framework}

%% file: figures/placeholder_teaser.tex
\begin{figure}[!t]
\centering
\includegraphics[width=\columnwidth]{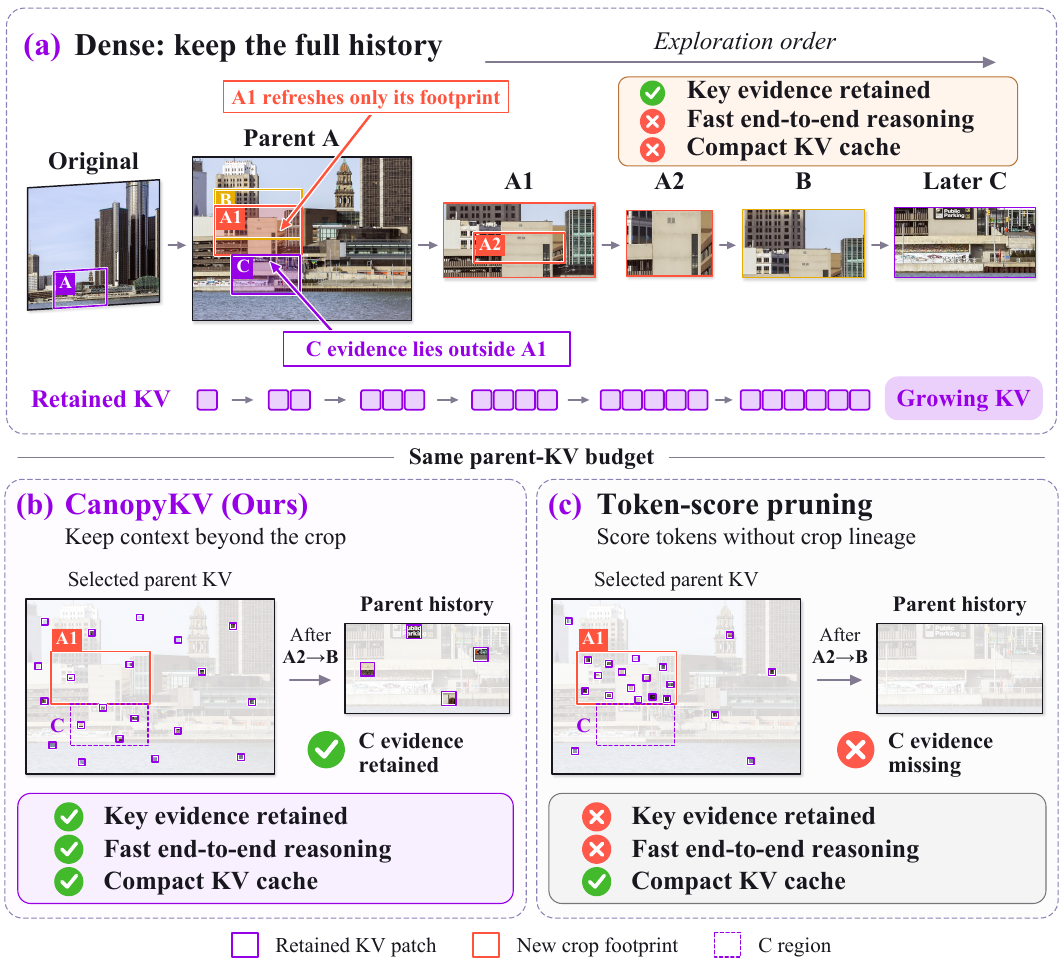}
\caption{\textbf{Local replacement and persistent complementarity.} (a) Dense retains all acquired views, growing its KV cache. A1 refreshes its footprint but leaves C outside it. (b,c) At equal parent-KV budgets, \method{} retains complementary C-region evidence across A2 and B, while token-score pruning may discard it. Insets show parent history, not a newly acquired C.}
\label{fig:teaser}
\end{figure}

%% file: figures/framework.tex
\begin{figure*}[t]
\centering
\includegraphics[width=\textwidth]{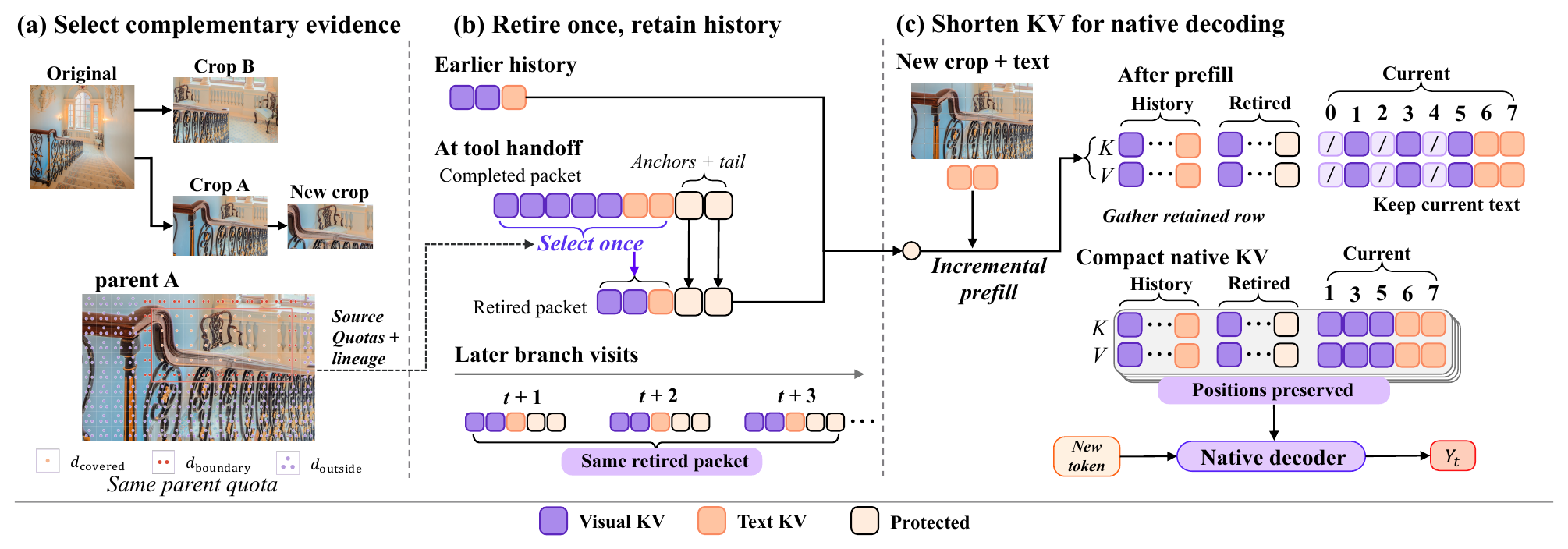}
\caption{\textbf{\method{} retains complementary evidence in compact KV.} (a) Source quotas and crop lineage favor context outside the incoming crop. (b) One-time retirement protects continuation state and fixes selected history across branch visits. (c) Incremental prefill is followed by visual pruning and KV compaction, preserving current text and logical positions.}
\label{fig:framework}\label{fig:architecture}\end{figure*}

%% file: sections/02_related_work.tex
\section{Related work}
\paragraph{Active visual reasoning.}
Visual programs compose perception tools~\cite{gupta2023visprog,suris2023vipergpt}, region inspection acquires targeted details~\cite{wu2023vstar,shao2024visualcot,zhao2025uvcot}, and Visual Sketchpad returns visual tool feedback~\cite{hu2024visualsketchpad}. Reinforcement learning trains interleaved reasoning and visual actions~\cite{zheng2025deepeyes,wang2025pixelreasoner,wu2026vtoolr1}, while Mini-o3 supports extended exploration~\cite{lai2025minio3}. \method{} manages the resulting history so that earlier evidence remains available at lower memory and execution cost.

\Needspace{5\baselineskip}
\paragraph{KV compression.}
General-purpose methods use attention statistics or sequence position~\cite{zhang2023h2o,xiao2023streaming}. Multimodal selectors incorporate text priority~\cite{wan2024lookm}, modality sparsity and cross-modal relevance~\cite{tu2025vlcache,huang2025aircache,liu2026tgvkv}, layerwise vision--text fusion~\cite{gu2026david}, semantic diversity~\cite{liu2026mixkv}, or query proxies~\cite{shu2026mmshift}. MuKV retrieves multi-grained video caches for later questions~\cite{xiao2026mukv}, while MetaCompress learns prompt-independent compression for unknown future questions~\cite{wang2026metacompress}. ActKV prioritizes action generation through attention-based retention~\cite{wang2026actkv}. \method{} uses executed visual actions to identify local replacement relationships, preserving complementary parent context that incoming observations leave uncovered. Future views here are acquired through the model's own actions, whereas MuKV and MetaCompress prepare memory for later questions. Each crop explicitly records which earlier region it revisits.

Visual token compression retains informative or diverse features~\cite{yang2025visionzip,alvar2025divprune}, combines temporal merging with dynamic KV selection~\cite{tao2025dycoke}, or progressively encodes complementary frame information~\cite{yang2025pvc}. Token Cropr learns a lightweight selector~\cite{bergner2025tokencropr}, while TopV optimizes once during prefill without attention scores~\cite{yang2025topv}, preserving FlashAttention compatibility~\cite{dao2022flashattention}. \method{} obtains retention signals from existing tool metadata and directly exposes compact KV tensors to incremental decoding.

\raggedbottom
\Needspace{5\baselineskip}
\paragraph{Tree structures and hierarchical compression.}
TCFormer clusters visual tokens, while ToMe merges similar tokens through bipartite matching~\cite{zeng2022tcformer,bolya2023tome}. FlashVID constructs trees for spatiotemporal merging~\cite{fan2026flashvid}, and GraceKV organizes KV entries and prototypes hierarchically~\cite{tian2026gracekv}. Our tree records executed tool calls: each edge links a crop to its parent, and its coordinates identify parent context beyond the new observation. This exploration structure guides retention for later reasoning.

%% file: sections/03_formulation.tex
\section{Method}
\label{sec:method}
\label{sec:formulation}
\method{} reduces the cost of visual exploration by preserving useful history in a form that is inexpensive to reuse. Source quotas and crop lineage select complementary evidence (\cref{fig:framework}a), stable continuation preserves its availability across branch visits (\cref{fig:framework}b), and native KV execution lowers the cost of using it (\cref{fig:framework}c).

\subsection{Problem setup}
\label{sec:problem-setup}
Active visual reasoning alternates generation with visual tool calls. Interaction $t\geq0$ produces a \emph{packet} $\mathcal P_t$ of visual inputs, associated text, generated reasoning, and control tokens. Packets begin with the original image at $t=0$ and at each subsequent tool handoff.

\paragraph{Current and historical state.}
\label{sec:current-budget}
We distinguish the current resident packet $\widetilde{\mathcal P}_t$, which supports the ongoing interaction, from retained history $\mathcal H_t$. At the next tool handoff, we compress the completed packet once, termed \emph{retirement}, and add its selected rows to history. We keep current text intact throughout the interaction and compact visual rows as described in \cref{sec:native-kv}.

Let $n_i=|\mathcal P_i|$ be the number of uncompressed KV rows in packet $i$, where $|\cdot|$ denotes row count. Dense retains all packets, giving $N_t^{\mathrm{dense}}$ rows and $M_t^{\mathrm{dense}}$ bytes of KV,
\begin{equation}
N_t^{\mathrm{dense}}=\sum_{i=0}^{t}n_i,
\qquad M_t^{\mathrm{dense}}=\beta N_t^{\mathrm{dense}},
\label{eq:dense-memory}
\end{equation}
where one row contains a token's K and V across all $L$ layers and $h_{\mathrm{kv}}$ KV heads. Its size is $\beta=2Lh_{\mathrm{kv}}d_hb$ bytes, with head dimension $d_h$ and $b$ bytes per element. For comparable packet sizes, the cache grows linearly with reasoning rounds, increasing the history processed during decoding (\cref{fig:teaser}a).

End-to-end time $T_{\mathrm{traj}}$ includes tool execution, visual encoding, incremental prefill, decoding, and cache maintenance. Writing these costs as $T_{\mathrm{tool}}$, $T_{\mathrm{encode}}$, $T_{\mathrm{prefill}}$, $T_{\mathrm{decode}}$, and $T_{\mathrm{maint}}$, respectively, gives
\begin{equation}
\begin{aligned}
T_{\mathrm{traj}}=\sum_t\bigl(&T_{\mathrm{tool}}+T_{\mathrm{encode}}+T_{\mathrm{prefill}}\\
&+T_{\mathrm{decode}}+T_{\mathrm{maint}}\bigr)_t.
\end{aligned}
\label{eq:trajectory-time}
\end{equation}
The sum covers all interactions until termination. Both subsequent observations and the stopping point depend on decisions made from retained history. Compression therefore changes per-step cost and can change the subsequent trajectory. Our objective is accurate task completion at lower total cost, which requires preserving useful evidence while keeping cache maintenance inexpensive. A retained view may support a later region comparison even when it is unused during the intervening steps. Evaluating only the immediate decoding cost would miss this delayed contribution, so we assess both complete trajectories and controlled evidence reuse.

\flushbottom
\subsection{What the next crop cannot replace}
\label{sec:evidence-observation}
We examine whether a full incoming crop can recover facts after their supporting parent evidence is removed, and whether recovery depends on the facts being inside or outside its footprint.

\input{figures/region_motivation}

The study covers 320 samples across VisualProbe, V*, HR-Bench 4K, HR-Bench 8K, and MME-Lite. We separately remove equal-sized parent-KV regions covering facts inside or outside the incoming crop, then supply the full new crop and read out each fact. This comparison tests which lost parent evidence the new observation can recover. All five groups show the same asymmetry (\cref{fig:region-motivation}). Deleting every row in the target mask lowers pooled fact-readout accuracy by 1.56 percentage points inside the crop, versus 24.84 points outside. Parent context beyond the new view is therefore a priority for retention.

Numbering observations in acquisition order, let $o_0$ be the root image and $o_s$ a crop of parent $o_{\pi(s)}$, where $\pi(s)<s$ is the parent index. These parent--child edges form the crop tree. Each source $s$ has footprint $R_s$ in normalized root-image coordinates. For a new descendant $o_u$, the tree identifies its ancestors and $R_u$ locates the region revisited within them.

These relationships guide selection toward parent context beyond the incoming crop. Once selected, that context must remain available across later branch visits. The next two sections implement structural selection and stable continuation.

%% file: figures/region_motivation.tex
\begin{figure}[t]
\centering
\includegraphics[width=\columnwidth]{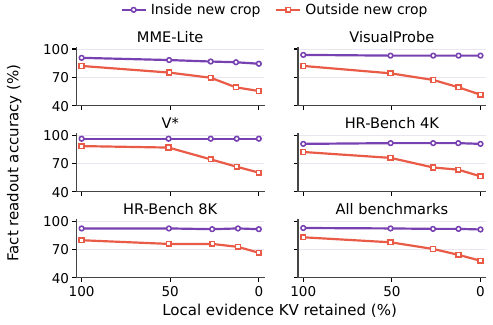}
\caption{\textbf{Parent evidence outside the new crop is harder to replace.} After removing matched amounts of parent KV inside or outside the incoming crop, we supply the full crop and read out the corresponding facts. The horizontal axes show KV retention within each target fact region.}
\label{fig:region-motivation}
\end{figure}

%% file: sections/04_method.tex
\subsection{Structural evidence selection}
Complementary evidence spans both observation sources and regions within a source. We allocate capacity among the root image, crop sources, and text, then use lineage to place each parent crop's quota where the new view leaves context uncovered. After caching the outgoing packet's remaining text, we protect its instruction and question tokens and last $w=32$ rows, forming set $\mathcal Q_t$. For retirement fraction $\rho\in(0,1]$, the retained row count $k_t$ is
\begin{equation}
k_t=\max\{|\mathcal Q_t|,\lceil\rho|\widetilde{\mathcal P}_t|\rceil\}.
\label{eq:packet-capacity}
\end{equation}
This retains all protected rows and leaves $k_t-|\mathcal Q_t|$ slots for the remaining candidates.

\paragraph{Source-aware quotas.}
\label{sec:source-quota}
Source quotas divide the retiring packet's budget among its available sources, while earlier history stays fixed. Unprotected visual rows form one group per source image, while unprotected text rows form one group. Each group $g$ receives an allocation weight $a_g$,
\begin{equation}
a_g=\begin{cases}a_{\mathrm{root}},&\text{root image},\\a_{\mathrm{crop}},&\text{each crop source},\\a_{\mathrm{text}},&\text{text}.\end{cases}
\label{eq:source-weights}
\end{equation}
The ordering $a_{\mathrm{crop}}>a_{\mathrm{text}}>a_{\mathrm{root}}>0$ prioritizes each crop source, followed by text and the root image. We allocate slots proportionally, cap quotas at available rows, and redistribute unused slots before uniform midpoint sampling.

Source quotas prevent a large source from dominating retention simply because it contributes more rows. Spatial allocation then directs each parent quota toward context beyond the new crop. Together, they preserve diversity across sources and complementary information within each parent.

\paragraph{Lineage-guided spatial retention.}
\label{sec:lineage}
Within each parent quota, lineage prioritizes context that the incoming crop cannot replace (\cref{sec:evidence-observation}). For a new crop $u$ with valid metadata, we follow its parent chain and adjust selections for ancestor crop sources $s>0$ in the retiring packet. Other branches keep their uniform selections.

The original patch indices and image grid map retained rows to cells in $R_s$. Comparing these cells with $R_u$ identifies fully covered cells, partially overlapping boundary cells, and cells outside the new crop. Their density weights satisfy
\begin{equation}
d_{\mathrm{outside}}>d_{\mathrm{boundary}}>d_{\mathrm{covered}}>0.
\label{eq:spatial-densities}
\end{equation}
Here each spatial class receives allocation weight equal to its density times its original candidate-row count. When capacity permits, each nonempty class first receives one representative, after which the remaining slots follow capped allocation and midpoint sampling. This preserves sparse representatives of revisited detail and favors context outside the new view (\cref{fig:framework}a). The source quota $q_s$, counting selected unprotected rows, remains fixed. Root-image, text, protected rows, and earlier history keep their selections.

\subsection{Stable continuation}
\label{sec:lifecycle}
At a handoff, protected rows $\mathcal Q_t$ and structurally selected rows form the retired packet (\cref{fig:framework}b). These rows remain fixed across later branch visits, keeping complementary evidence available as search focus changes.

Alongside fixed evidence, protected continuation state connects the next observation to the ongoing task. Instruction and question tokens preserve the task specification, while the protected tail retains recent source-name tokens and message delimiters. Protected rows count toward $k_t$, with overlaps counted once. The retirement budget thus covers both complementary evidence and the state needed to continue reasoning from it.

Let $\widetilde n_t=|\widetilde{\mathcal P}_t|$ be the current packet's resident row count. Combining it with the retained counts $k_i$ of earlier packets gives the resident cache $\mathcal C_t$,
\begin{equation}
\mathcal C_t=\mathcal H_t\cup\widetilde{\mathcal P}_t,
\qquad |\mathcal C_t|=\sum_{i<t}k_i+\widetilde n_t.
\label{eq:resident-memory}
\end{equation}
Retirement separates selection from later reuse: each packet is selected once, and its retained evidence remains available without predicting the order of future branch visits. Earlier selections are gathered and reused rather than ranked again.

\input{tables/01_quality}
\subsection{Efficient native KV execution}
\label{sec:native-kv}
\method{} converts tool-provided parent relations and coordinates into selections for compact native KV tensors. At a new-image handoff, we gather the completed packet's selected K and V with earlier history chronologically at every layer. Incremental prefill processes appended tool text and the full new image using this prefix (\cref{fig:framework}c).

To match the target mean total-KV budget $B$ (\cref{sec:experiments}), we calibrate the current-image retention fraction $\alpha\in(0,1]$ using measured physical KV, with retirement fraction $\rho$ fixed. After full prefill, uniform midpoint sampling retains $\max\{1,\lceil\alpha n_v\rceil\}$ of the new image's $n_v$ visual rows, and we gather their KV. Current text stays intact, and retained visual rows preserve their original patch indices for later spatial selection.

The indices in \cref{fig:framework}c denote original token positions. Gathering preserves stored values and encoded positional information, while a separate row map retains the original token positions. After compaction, the decoder reads a shorter KV tensor at each subsequent step. Each image is encoded once, and generation reuses the retained cache while appending new token KV.

\paragraph{Low-cost cache management.}
At handoffs, historical packets undergo structural selection at retirement, while new images undergo uniform visual retention after full prefill. Neither requires attention extraction or auxiliary forward passes. Subsequent decoding reuses compact KV without reselection. Maintenance comprises metadata processing and tensor gathering. Gathering $m_t$ retained rows at interaction $t$ moves $\beta m_t$ bytes, including fixed history, and reduces the cache read during subsequent decoding.

The full algorithm and allocation details appear in Appendix~\ref{supp:lifecycle}, with default weights in Appendix~\ref{supp:runtime}.

%% file: tables/01_quality.tex
\begin{table*}[t]
\caption{\textbf{Answer accuracy and reasoning speed.} Acc. is in percent, and Speed is relative to Dense ($\times$). Bold marks column maxima. Compressed methods target $B\approx0.25M_{\mathrm{Dense}}$ total KV. Actual total-KV retention spans 24.07--26.84\% of Dense, except LOOK-M (30.97\%) and AirCache (30.18\%) on MME-Lite. Full per-dataset retentions are listed in Appendix~\ref{supp:current-budget}, \cref{tab:current-budget}.}
\label{tab:quality}\label{tab:efficiency}
\centering\small
\setlength{\tabcolsep}{3pt}\renewcommand{\arraystretch}{1.12}
\begin{tabularx}{\textwidth}{ll*{10}{>{\centering\arraybackslash}X}}
\toprule
\multirow{2}{*}{\textbf{Method}} & \multirow{2}{*}{\textbf{Venue}} & \multicolumn{2}{c}{\textbf{VisualProbe}} & \multicolumn{2}{c}{\textbf{V*}} & \multicolumn{2}{c}{\textbf{HR-Bench 4K}} & \multicolumn{2}{c}{\textbf{HR-Bench 8K}} & \multicolumn{2}{c}{\textbf{MME-Lite}}\\
\cmidrule(lr){3-4}\cmidrule(lr){5-6}\cmidrule(lr){7-8}\cmidrule(lr){9-10}\cmidrule(lr){11-12}
& & Acc.$\uparrow$ & Speed$\uparrow$ & Acc.$\uparrow$ & Speed$\uparrow$ & Acc.$\uparrow$ & Speed$\uparrow$ & Acc.$\uparrow$ & Speed$\uparrow$ & Acc.$\uparrow$ & Speed$\uparrow$\\
\midrule
Dense & \tablevenue{Reference} & 51.26 & 1.00 & 82.20 & 1.00 & 72.38 & 1.00 & 67.38 & 1.00 & 43.04 & 1.00\\ 
\midrule
H$_2$O~\cite{zhang2023h2o} & \tablevenue{NeurIPS\,'23} & 50.10 & 0.38 & 31.94 & 0.10 & 66.25 & 0.30 & 66.88 & 0.37 & 42.68 & 0.44\\ 
LOOK-M~\cite{wan2024lookm} & \tablevenue{EMNLP-F\,'24} & 46.41 & 0.55 & 81.15 & 0.53 & 67.75 & 0.51 & 64.38 & 0.53 & 39.97 & 0.74\\ 
VL-Cache~\cite{tu2025vlcache} & \tablevenue{ICLR\,'25} & 53.59 & 1.18 & 81.68 & 1.04 & 70.75 & 0.91 & 67.63 & 0.97 & 46.27 & 1.25\\ 
AirCache~\cite{huang2025aircache} & \tablevenue{ICCV\,'25} & 46.21 & 0.83 & 78.53 & 0.58 & 71.00 & 0.89 & 64.25 & 0.90 & 37.99 & 0.93\\ 
MixKV~\cite{liu2026mixkv} & \tablevenue{ICLR\,'26} & 51.84 & 0.93 & 83.25 & 0.75 & 74.00 & 0.89 & 68.13 & 0.97 & 49.82 & 1.11\\ 
TGV-KV~\cite{liu2026tgvkv} & \tablevenue{ICML\,'26} & 46.60 & 0.82 & 72.77 & 0.59 & 67.63 & 0.69 & 65.75 & 0.81 & 41.79 & 0.98\\ 
MM-ShiftKV~\cite{shu2026mmshift} & \tablevenue{ACL-F\,'26} & 49.90 & 0.61 & 79.58 & 0.48 & 70.00 & 0.61 & 66.13 & 0.65 & 42.21 & 0.76\\ 
AttentionPack~\cite{ilhan2026attentionpack} & \tablevenue{CVPR\,'26} & 34.37 & 0.26 & 79.06 & 0.32 & 65.75 & 0.32 & 62.50 & 0.35 & 39.60 & 0.41\\ 
\addlinespace[2pt]\rowcolor{avgreen!9}
\method{} (ours) & \tablevenue{This work} & \textbf{55.53} & \textbf{1.31} & \textbf{85.34} & \textbf{1.30} & \textbf{75.63} & \textbf{1.13} & \textbf{71.13} & \textbf{1.19} & \textbf{53.52} & \textbf{1.64}\\ 
\bottomrule\end{tabularx}\end{table*}

%% file: sections/05_experiments.tex
\section{Experiments}
\label{sec:experiments}
We test whether compact history supports accurate task completion at lower cost, why its contents and lifetime matter, and whether execution savings persist when the reasoning stream is fixed.

\subsection{Experimental setup}
\paragraph{Models and benchmarks.}
We evaluate Mini-o3-7B-v1~\cite{lai2025minio3}, based on Qwen2.5-VL~\cite{bai2025qwen}, on the full sets of VisualProbe~\cite{lai2025minio3} (515 questions), V*~\cite{wu2023vstar} (191), HR-Bench~\cite{wang2024hrbench} (800 each for 4K/8K), and MME-RealWorld Lite~\cite{zhang2025mmerealworld} (1,919, abbreviated MME-Lite). Methods share ordinary reasoning prompts, tools, and scorers, including Qwen2.5-72B-Instruct~\cite{qwen2024qwen25} for VisualProbe. We use BF16 greedy decoding with one trajectory per question, capped at 32 rounds, 8,192 generated tokens/round, and 32,768 context tokens.

\paragraph{Comparison protocol.}
Methods run incrementally on H20 GPUs (batch size one), using published selectors and available official efficiency kernels. The target mean total physical KV is $B\approx0.25M_{\mathrm{Dense}}$, with $M_{\mathrm{Dense}}$ denoting the dataset's mean Dense KV. The budget counts visual/text K/V tensors across all physically stored heads. Timing covers encoding, generation, cache management, and tool execution. Additional experimental settings are provided in Appendix~\ref{supp:runtime}.

\subsection{Accuracy and reasoning speed}
\label{sec:matchedbudget}
\paragraph{Compact history improves accuracy and speed.}
\method{} achieves the best reported accuracy in every group of \cref{tab:quality} and exceeds Dense in all five. On VisualProbe, it reaches 55.53\%, versus 51.26\% for Dense and 53.59\% for VL-Cache, with an 8.49-point gain over Dense on Hard (\cref{tab:main-difficulty}). Retaining the full history is therefore unnecessary for accurate active visual reasoning.

On MME-Lite, \method{} retains 24.63\% of total KV and achieves 53.52\% accuracy, exceeding Dense by 10.47 points and the strongest baseline, MixKV, by 3.70 points. This removes 75.37\% of resident KV while preserving useful evidence. Complete-task time falls from Dense's 53.12 to 32.35 seconds per question, a $1.64\times$ speedup, while rounds fall from 10.01 to 6.04. Across all five groups, \method{} is faster than every compared baseline, with speedups over Dense of $1.13$--$1.64\times$ (\cref{tab:quality}).

\par\vspace{3.25ex}
\noindent\textbf{Compression must sustain visual search.}
Compression must preserve evidence for later decisions while reducing the cost of using it. The main comparison shows that smaller caches do not necessarily improve accuracy or completion time. Budget sweeps reveal increased search effort with insufficient evidence (\cref{fig:budget-sweep}), while fixed-trajectory measurements expose cache-maintenance overhead (\cref{sec:execution}). Effective compression therefore requires both useful retained content and efficient execution.

The objective is accurate search at lower cost, so fewer rounds alone do not establish that benefit. MixKV uses 5.32 rounds on MME-Lite but takes 47.66 seconds, compared with 32.35 for \method{}. On V*, \method{} takes slightly more rounds than Dense (4.02 versus 3.80), yet finishes $1.30\times$ faster (\cref{tab:main-rounds}). Search depth and execution cost therefore need to be understood together.

\subsection{Compression and search effort}
\label{sec:budget-sweep}
\input{figures/budget_sweep}
The budget sweep tests whether reducing memory also reduces the work needed to finish a task. A smaller cache lowers per-step work, but retained evidence affects subsequent search. \Cref{fig:budget-sweep} examines this interaction across KV budgets on a fixed subset of 358 VisualProbe questions.

\method{} preserves useful search over a broad range of budgets. Across six settings retaining 24.28--94.06\% KV, it improves both accuracy and completion time over Dense. At 34.84\% retention, it reaches 49.44\% accuracy in 26.29 seconds, compared with Dense's 43.85\% and 39.86 seconds. Stronger pruning eventually reverses the benefit: at 12.65\%, accuracy falls to 35.75\%, search expands to 11.37 rounds, and time rises to 40.15 seconds.

Longer search also offsets memory savings in the baselines. Reducing LOOK-M retention from 45.92\% to 23.36\% increases rounds from 10.26 to 12.87 and time from 68.09 to 76.59 seconds. AirCache likewise requires more work at its strongest compression. These curves show that memory savings improve efficiency while retained evidence sustains effective continuation. The retention ranges of the plotted methods overlap at 23.36--92.29\% KV.

\subsection{Component ablation}
\label{sec:component-ablation}
\input{tables/20_four_component_ablation}
The four removals in \cref{tab:four-component-ablation} assess the roles of evidence content, persistence, and executable continuation under a common target KV budget. All variants are evaluated on the complete VisualProbe set. Removing crop lineage preserves the crop tools and source quotas but selects uniformly within each source.

\paragraph{Selecting complementary evidence.}
Removing source quotas lowers accuracy from 55.53\% to 53.79\%. Removing crop lineage lowers accuracy to 46.99\%, while mean rounds remain close to the full method at 7.14 versus 7.18. This separation between answer quality and search length highlights the importance of which evidence occupies the cache. Source allocation and lineage jointly determine that content.

\paragraph{Keeping evidence and handoff state available.}
Removing fixed history lowers accuracy to 52.23\%. Removing tail protection lowers it to 47.38\% and increases rounds to 11.97. The latter intervention both damages answer quality and expands search, showing the role of the recent handoff state in continuation. These results motivate the following matched-capacity studies of what is selected, what survives later branch visits, and what remains usable when reasoning resumes.

\subsection{Preserving evidence across search}
\label{sec:mechanism}
\input{figures/placeholder_growth}
\paragraph{Crop lineage locates complementary evidence.}
At the same capacity, lineage preserves more of the parent information outside the incoming view. We test this through delayed readout after visits to other branches. In synthetic scenes, we compare lineage-guided retention with uniform selection at matched source quotas and KV capacity. The comparison isolates where the parent evidence is retained. Correct lineage retains 46.25\% of unprotected rows outside the new crop, versus 23.83\% for the control, and improves readout accuracy for this region by 23.63--29.49 points to 93.16--95.90\% across branch visits (\cref{fig:growth}a).

\looseness=-1
The same selection principle transfers to natural-image questions. Following the same crop sequences at matched KV capacity, crop lineage improves delayed four-option readout accuracy by 4.69--5.47 points after 0, 8, or 16 visits to other branches (\cref{fig:growth}b). These readouts test whether retained evidence remains usable after exploration. Matching the crop sequence holds acquired information constant, while delaying the question separates retention from immediate query relevance. The improvement therefore supports the intended role of complementary history: carrying information across intervening observations until a later decision requires it.

The corresponding real-image selections show how this capacity shifts away from repeated coverage (\cref{fig:equal-parent-case}).

\paragraph{Fixed history preserves evidence until reuse.}
Useful selections must survive later handoffs. Starting from identical caches at matched capacity, \cref{fig:persistence} compares keeping retired selections fixed with resampling them uniformly. After eight visits to other branches, resampling reduces retained parent rows from 296 to 6 and readout accuracy to 22.46\%, whereas fixed history maintains 90.82\% accuracy. Repeated selection therefore discards useful evidence even when capacity is unchanged. On natural-image questions, fixed history also improves readout accuracy from 50.98\% to 57.81\% after 16 visits.

Together, selection and persistence preserve context that the current crop cannot supply: lineage determines what enters history, and fixed retention carries it through later exploration. Handoff protection preserves the state needed to use that history.

\input{figures/analysis_persistence}
\paragraph{Handoff protection sustains continuation.}
Preserving the recent handoff state helps the model use its retained history. On natural-image confirmation questions, replacing the protected 32-row tail at equal capacity lowers accuracy from 72.40\% to 65.63\% (\cref{tab:ablation}) and increases remaining rounds from 3.38 to 4.82. The protected region contains recent source-name tokens and message delimiters. Keeping this state supports accurate continuation and reduces the extra interactions observed after its removal.

\subsection{Low-cost cache execution}
\label{sec:execution}
To isolate execution benefits from changes in search, we replay eight long MME-Lite sequences, giving every method the same observations and continuation tokens. Fixing the reasoning stream exposes the cost of maintaining and using the retained cache.

\looseness=-1
For \method{}, selection and compaction take 0.11 seconds per sequence. Including cache append and layout operations, cache work totals 3.72 seconds, versus 11.91 for VL-Cache and 22.83 for MixKV (\cref{fig:cache-update-runtime}a). Tool metadata supplies retention signals without model-derived importance statistics.

Physical cache size and maintenance cost therefore matter jointly. Fewer retained rows reduce attention work, but execution gains depend on whether the savings outweigh selection and layout costs. Fixed-stream replay isolates this balance from changes in search.

Selection at image handoffs and fixed historical choices amortize maintenance over later decoding steps. Between handoffs, native attention directly reads the compact tensors, reducing execution cost without reranking earlier evidence.

\input{figures/structural_execution_cost}

\begingroup\widowpenalty=10000
\method{} is the fastest of all ten displayed methods on identical reasoning streams: 66.5 seconds versus Dense's 92.3, a $1.39\times$ execution speedup (\cref{fig:cache-update-runtime}b). On these long streams, the speedup therefore persists without reducing the number of observations or continuation tokens. The $1.64\times$ complete-task MME-Lite speedup additionally reflects freely generated search and tool use.

These experiments connect memory reduction to effective exploration: complementary evidence supports later decisions, persistent retention keeps it available, and compact native execution lowers the cost of using it. Additional mechanism studies, execution protocols, and extended results are collected in Appendices~\ref{supp:figure-specifications}, \ref{supp:execution-diagnostics}, and~\ref{supp:current-budget}.\par
\endgroup

%% file: figures/budget_sweep.tex
\begin{figure}[t]
\centering
\includegraphics[trim=0 9bp 0 1bp,clip,width=\columnwidth]{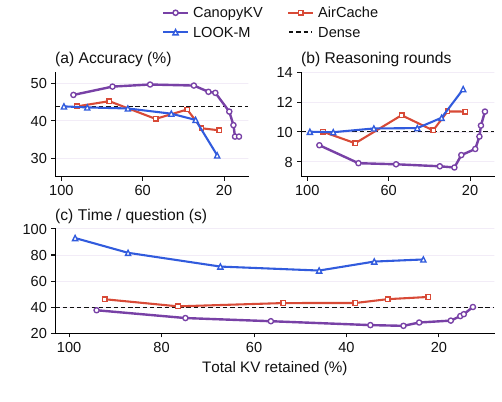}
\caption{\textbf{Compression changes both memory and search effort.} Budget sweep on the fixed VisualProbe subset. Dashed lines denote Dense.}
\label{fig:budget-sweep}
\end{figure}

%% file: tables/20_four_component_ablation.tex
\begin{table}[!ht]
\caption{\textbf{Single-component ablation on VisualProbe.} Each variant removes one component. KV is normalized to Dense.}
\label{tab:four-component-ablation}
\centering\small
\setlength{\tabcolsep}{3pt}\renewcommand{\arraystretch}{1.12}
\begin{tabularx}{\columnwidth}{@{}l*{4}{>{\centering\arraybackslash}X}@{}}
\toprule
Configuration & Acc. & Total KV & Rounds & Time\\
 & (\%) $\uparrow$ & (\%) $\downarrow$ & & (s) $\downarrow$\\
\midrule
Dense & 51.26 & 100.00 & 7.84 & 30.43\\
\method{} (full) & \textbf{55.53} & 24.34 & 7.18 & 23.23\\
\midrule
w/o source quotas & 53.79 & 26.43 & 7.10 & 26.44\\
w/o crop lineage & 46.99 & 26.22 & 7.14 & 22.45\\
w/o fixed history & 52.23 & 24.10 & 7.27 & 24.81\\
w/o tail protection & 47.38 & 26.22 & 11.97 & 38.10\\
\bottomrule
\end{tabularx}
\end{table}

%% file: figures/placeholder_growth.tex
\begin{figure}[!ht]
\centering
\includegraphics[width=\columnwidth]{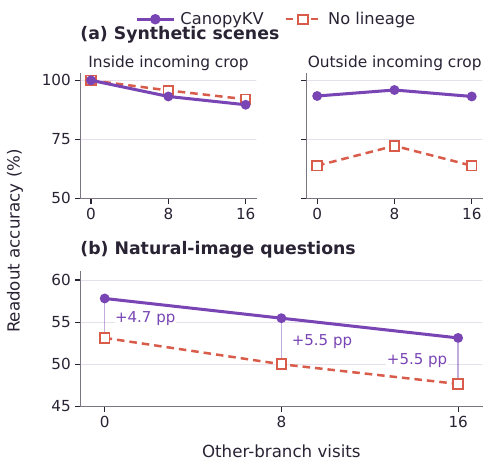}
\caption{\textbf{Crop lineage preserves complementary evidence.} Delayed readout at matched source quotas and KV capacity for (a) synthetic facts inside/outside the incoming crop and (b) natural-image questions.}
\label{fig:growth}
\end{figure}

%% file: figures/analysis_persistence.tex
\begin{figure}[tbp]
\centering
\includegraphics[width=\columnwidth]{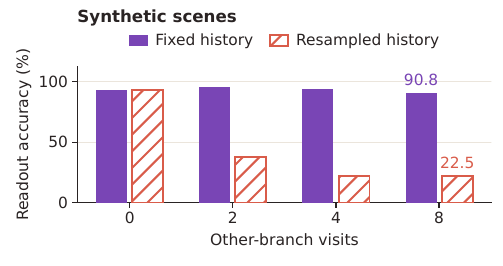}
\caption{\textbf{Stable history survives branch changes.} Fixed versus uniformly resampled history on 128 synthetic scenes at matched KV capacity.}
\label{fig:persistence}
\end{figure}

%% file: figures/structural_execution_cost.tex
\begin{figure}[t]
\centering
\includegraphics[width=\columnwidth,trim=0 8.5bp 0 3.5bp,clip]{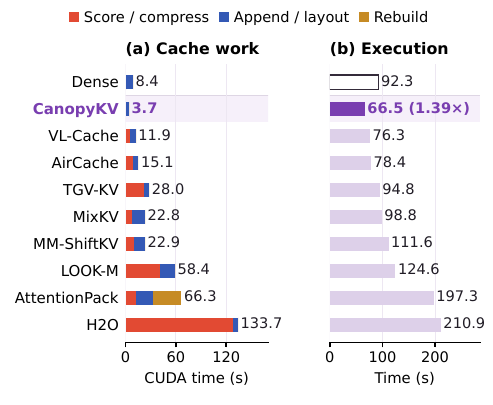}
\caption{\textbf{Low cache overhead and fast execution.} Cache-operation costs (a) and execution time (b) on identical long MME-Lite search streams.}
\label{fig:cache-update-runtime}
\end{figure}

%% file: sections/06_conclusion.tex
\Needspace{6\baselineskip}
\section{Conclusion}
\label{sec:conclusion}
Effective compression for active visual reasoning must preserve evidence for future exploration while reducing the cost of reusing it. \method{} turns local evidence replacement and persistent complementarity into a training-free framework that selects complementary history, keeps it stable across branches, and protects the state needed for continued reasoning. Tool metadata and compact native KV jointly reduce memory-maintenance overhead and subsequent cache-read costs. Across four benchmarks, \method{} improves both answer accuracy and reasoning speed over the compared baselines. These results show that jointly designing for persistent evidence availability and low-cost reuse supports more accurate and efficient visual exploration.

%% file: sections/07_supplement.tex
\paragraph{Guide.}
The supplement provides implementation details, settings, mechanism studies, execution analysis, qualitative illustrations, complete results, and cache growth in Appendices~A--G, respectively.

\section{Implementation details}
\label{supp:lifecycle}
\subsection{Retirement and positions}
Native compaction preserves selected K/V values, key rotations, and original logical positions~\cite{bai2025qwen}. All layers and native KV heads share the selected row indices, and new positions follow the full transcript. Anchors include the initial instruction/question and nonvisual image delimiters. A packet shorter than its protected tail remains complete. Retiring a packet with capacity $k$ leaves $|\mathcal H|+k$ rows, where $\mathcal H$ is earlier history. The terminal packet remains resident during answer generation.

\Cref{alg:handoff} specifies the handoff order. Current-visual compaction follows full prefill, so its peak includes the incoming image's complete KV.

\begin{algorithm}[!ht]
\caption{Replacement-aware retirement at a tool handoff}
\label{alg:handoff}
\begin{algorithmic}[1]
\raggedright
\Require Native cache $\mathcal H\cup\widetilde{\mathcal P}$; source tree; tool reply
\State Commit the outgoing segment's remaining text
\State Protect anchors and tail; compute $k$ by \cref{eq:packet-capacity}
\State Allocate source quotas; choose spaced representatives
\For{crop source $s>0$ in the completed packet}
\If{fresh crop $u$ has valid metadata and $s$ is its \mbox{strict ancestor}}
\State \parbox[t]{\dimexpr\linewidth-\algorithmicindent-\algorithmicindent\relax}{\raggedright Classify $s$'s patches against the \mbox{incoming footprint}}
\State Redistribute its fixed quota using \cref{eq:spatial-densities}
\EndIf
\EndFor
\State Gather selected native KV; freeze the retired selection
\State Append the new tool suffix and observation
\State Apply the configured current-visual budget after prefill
\State Continue native decoding
\end{algorithmic}
\end{algorithm}

\subsection{Quota allocation and spatial selection}
\label{supp:allocation}
For optional budget $B_{\mathrm{opt}}=k-|\mathcal Q|$, each group $g$ has $c_g$ candidates, weight $a_g$, and initial allocation $q_g=0$. With remaining budget $B_{\mathrm{rem}}$ and unsaturated groups $U$, allocate
\begin{equation}
v_g=B_{\mathrm{rem}}\frac{a_g}{\sum_{j\in U}a_j},\qquad
\Delta_g=\min\{c_g-q_g,\lfloor v_g\rfloor\}.
\label{eq:capped-allocation}
\end{equation}
Repeat after updating allocations and remaining capacity. If all increments vanish, distribute rows in descending $v_g$ order, breaking ties by group index. Source weights follow main-paper \cref{eq:source-weights}. From chronologically ordered candidates $r_0,\ldots,r_{c-1}$, a quota $q>0$ selects
\begin{equation}
\operatorname{Space}(r,q)=\{r_{\lfloor(2j+1)c/(2q)\rfloor}:0\leq j<q\}.
\label{eq:spaced}
\end{equation}

Spatial redistribution applies only to strict crop ancestors with valid metadata. Original patch footprints are mapped into root-image coordinates and classified as covered, intersecting the boundary, or outside the new crop. Protected rows remain selected. When capacity permits, each nonempty class receives one row before remaining slots are allocated in proportion to its original candidate-row count times the corresponding density in main-paper \cref{eq:spatial-densities}. Single-class sources and unreliable grids keep their base selection, as do the root and other branches.

\section{Experimental settings}
\label{supp:runtime}
\subsection{Evaluation and measurement}
\label{supp:runtime-measurement}
Shared settings follow main-paper Section~4.1. Image pixel limits are 40,000--2,000,000. Failed or unanswered questions count as incorrect, and MME-Lite uses its released choice parser. All methods follow the ordinary evaluation loop.

CUDA-synchronized request timing includes encoding, incremental inference, cache management, and tool execution, excluding loading, separate warmup, and offline scoring. Speedup divides Dense's mean time by the method's.

\paragraph{Physical KV.}
\label{supp:memory-accounting}
We count resident K+V across all layers and native heads, including current/historical visual rows and text. Values are averaged over post-prefill/compaction boundaries within each question, then over questions. This excludes weights, activations, workspaces, and transient coexistence of old and replacement tensors during gathering.

\paragraph{Budget settings.}
The main comparison and component removals share a fixed retirement fraction $\rho$. For each dataset and variant, we calibrate current-image retention $\alpha$ from measured physical KV to match the target mean total-KV budget. MME-Lite uses stratified calibration and a disjoint confirmation subset for LOOK-M. Calibration questions remain in the full evaluation. Actual retention is reported with the corresponding results (\cref{tab:current-budget,tab:four-component-ablation}).

\paragraph{Allocation defaults.}
\label{supp:allocation-defaults}
We use source weights $(a_{\mathrm{root}},a_{\mathrm{crop}},a_{\mathrm{text}})=(1,4,2)$ and spatial densities $(d_{\mathrm{covered}},d_{\mathrm{boundary}},d_{\mathrm{outside}})=(1,2,3)$.

\paragraph{Budget sweep.}
\label{supp:budget-protocol}
We form a method-independent eligible subset containing all 358 VisualProbe questions whose Dense trajectories executed at least three crops. The same eligibility rule and question list define the evaluation subset for every budget setting, including failed trajectories. Horizontal coordinates divide same-subset mean total KV by Dense's mean.

\subsection{Baseline adaptations}
\label{supp:baseline-adaptation}
Implementations preserve published selectors, update boundaries, and multimodal positions with four native KV heads. MixKV combines the released AdaKV selector~\cite{feng2025adakv}, online head-wise mixing, CUDA append, and variable-length FlashAttention-2~\cite{dao2024flashattention2}, using SDPA for rectangular layouts. MM-ShiftKV uses its released query-proxy selector, CUDA append, and FlashAttention. VL-Cache uses Triton importance statistics.

H2O averages scores over each native KV head's query group and evicts after input chunks and decoding steps. AttentionPack compresses before rotary encoding and restores original positions after full-KV reconstruction ($F=1$). Its VisualProbe visual K/V ranks are 67, text-K rank is 64, and text V is uncompressed, with a 100-row SVD buffer threshold.

\section{Supplementary mechanism studies}
\label{supp:figure-specifications}
\label{supp:source-priors}
\label{supp:mme-mechanisms}
\subsection{Lineage and handoff protection}
\Cref{tab:ablation} summarizes equal-capacity interventions. In the lineage tests, questions are revealed only at readout, and caches match total, visual, and text capacity.

\input{tables/03_ablation}

\paragraph{Natural-image lineage.}
\label{supp:lineage-transfer}
A separate 64-question, 62-image-cluster study fixes real crop streams and evaluates two answer orders. At $\rho=0.3206266$, question-averaged lineage accuracy is 57.81/55.47/53.12\% after 0/8/16 visits, versus 53.12/50.00/47.66\% without lineage.

\subsection{Spatial replacement and selection}
\paragraph{Local replacement.}
\label{supp:local-replacement}
Main-paper \cref{fig:region-motivation} uses 64 samples from each of the five benchmark groups, totaling 320 samples in 290 source-image clusters. Selection checks fact visibility, crop geometry, text prefixes, and native budgets. The 96 development samples are excluded. At a saved Dense handoff, covered and outside fact masks are matched in KV rows. Deletion precedes full prefill of the incoming crop, preserving all text, retained KV values, and positions. Local facts are read in two answer orders, averaged within facts and then equally over benchmark samples.

The plotted condition removes other historical visual sources and uses a real incoming crop. Target-mask retention is $q\in\{1,0.5,0.25,0.1,0\}$, plotted at actual integer-row fractions. Zero deletes the target mask, not the whole parent. These local fact-readout scores are separate from original-question accuracy.

\paragraph{Repeated coverage.}
\label{supp:mme-selection}
Sixteen MME-Lite prefixes defined by geometry and hash compare AirCache, MM-ShiftKV, TGV-KV, \method{}, and uniform midpoint sampling on the same Dense parent candidates and original \method{} quotas. Results average all saved heads at layer~15, then cases equally.

\begin{figure}[!t]
\centering
\includegraphics[width=\columnwidth]{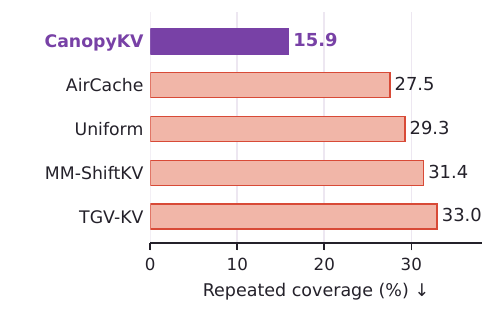}
\caption{\textbf{Lineage reduces repeated coverage across MME crops.} Mean fraction of selected parent tokens overlapping the incoming crop, using equal parent quotas and averaging all saved heads.}
\label{fig:mme-coverage}
\end{figure}

\method{} assigns 15.90\% of retained parent tokens to the new footprint, versus 27.54--32.97\% for scored selectors and 29.26\% for uniform sampling. Its head-averaged overlap is lower in all sixteen cases (\cref{fig:mme-coverage}).

\subsection{Operating conditions}
\label{supp:current-evidence}
The handoff study uses 52 discovery and 192 distinct-image confirmation cases. \Cref{tab:ablation} reports equal-capacity replacement of the protected 32-row tail, which contains source-name suffixes and closing syntax.

%% file: tables/03_ablation.tex
\begin{table}[!ht]
\caption{\textbf{Lineage and handoff protection at equal cache capacity.} Readout accuracy over 128 synthetic scenes and answer accuracy over 192 natural-image questions.}
\label{tab:ablation}
\centering\small\setlength{\tabcolsep}{3pt}
\begin{tabular}{lrrr}\toprule
Intervention & Cases & \shortstack{Full\\(\%)} & \shortstack{Control\\(\%)}\\
\midrule
Lineage (both regions, 16 visits) & 128 & 91.41 & 77.93\\
Protected handoff tail & 192 & 72.40 & 65.63\\
\bottomrule\end{tabular}
\end{table}

%% file: sections/09_analysis_archive.tex
\section{Execution analysis}
\label{supp:execution-diagnostics}
\subsection{Fixed-input replay}
\label{supp:operation-timing}
Main-paper \cref{fig:cache-update-runtime} uses eight MME-Lite streams with sixteen crops and at least one nonroot parent, selected by domain rotation and image-deduplicated hash order. All ten methods receive identical pixels, crop geometry, text, and evaluation inputs. KV is averaged across seventeen boundaries and streams. \method{} uses $\rho=0.3206266$, $\alpha=0.4569659$. Baseline capacities match its mean residency, except AirCache and LOOK-M retain their native settings. Compressed retention spans 19.99--22.14\% of Dense, with \method{} at 21.01\%.

On H20 GPUs, each method completes a full warmup and three repetitions with rotated order. CUDA-synchronized segments include encoding, incremental inference, cache work, and production bookkeeping. They exclude disk I/O, externally fixed tool execution, sampling, and result writing. Separate profiling passes measure elapsed cache-operation intervals, including host-launch gaps, independently of total replay time. Scoring/compression and reconstruction are separate categories.

\subsection{Generated trajectories}
\label{supp:operation-timing-generated}
To quantify cache growth during exploration (\cref{fig:exploration-growth}), we follow 112 Dense VisualProbe questions reaching six crops, using the first generation call at each boundary. Mean post-prefill KV grows from 150.74 to 606.31 MiB ($4.02\times$). Per-token backend time is normalized to each question's original-image call before taking the median. Its interval includes sampling, cache updates, and metrics.

\Cref{fig:trajectory-backend} uses the complete VisualProbe set and the timing boundaries in Appendix~\ref{supp:runtime-measurement}. Prefill includes image encoding, append, and some maintenance. Backend includes attention, sampling, instrumentation, and cache updates. Per-token means divide accumulated backend time by generated tokens.
\begin{figure}[!ht]
\centering
\includegraphics[width=\columnwidth]{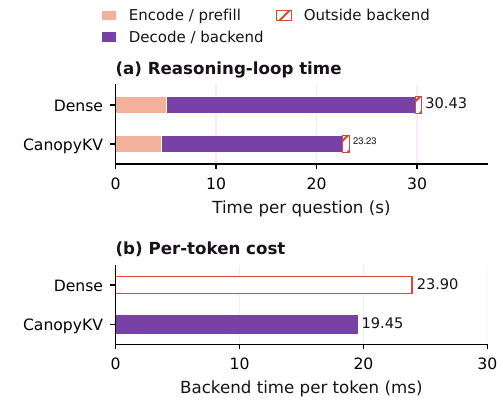}
\caption{\textbf{Compact history reduces trajectory and per-token cost.} Top: request-time breakdown. Bottom: backend time per generated token. Each method uses the complete VisualProbe evaluation set.}
\label{fig:trajectory-backend}\end{figure}

%% file: sections/14_selection_gallery.tex
\section{Qualitative illustrations}
\label{supp:selection-gallery}
\subsection{Spatial-selection visualization}
\label{supp:equal-parent-selection}
\input{figures/appendix_equal_parent_case}
This figure visualizes spatial selections and is provided for qualitative interpretation rather than quantitative evaluation. Scored baselines use their original scorers on shared Dense Q/K; all methods select from the same resident parent candidates. In \cref{fig:equal-parent-case}, each method retains 28 tokens, representing 10.04\% of 279 candidates and 5.02\% of the original $18\times31$ grid. Full-color patches denote retained tokens, and overlap includes boundary intersections.

%% file: figures/appendix_equal_parent_case.tex
\begin{figure}[!ht]
\centering
\includegraphics[trim=2.5bp 0 2.5bp 0,clip,width=\columnwidth]{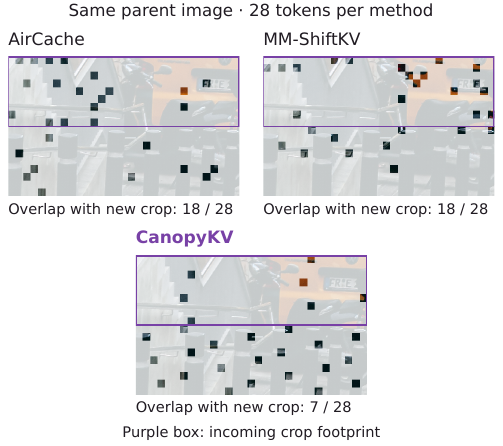}
\caption{\textbf{Spatial selections at an equal parent-token budget.} Each method retains 28 parent tokens. Purple marks the incoming crop.}
\label{fig:equal-parent-case}
\end{figure}

%% file: sections/08_additional_experiments.tex
\FloatBarrier
\section{Complete results}
\label{supp:current-budget}
\subsection{Main-comparison statistics}
\label{supp:main-comparison-details}
\Cref{tab:current-budget,tab:main-difficulty,tab:main-rounds} give actual KV, difficulty accuracy, and rounds. KV uses the definition in Appendix~\ref{supp:memory-accounting}. Reused Full entries share these values. MME-Lite quality and speed use the same complete trajectories.

\input{tables/16_current_budget}

\input{tables/22_main_difficulty}

\input{tables/23_main_rounds}

\subsection{Low-budget extensions}
\label{supp:low-budget-extension}
Four AirCache~\cite{huang2025aircache} and LOOK-M~\cite{wan2024lookm} settings target 14.63\% and 12.65\% total KV on the same 358-question sweep subset. Their realized retention is higher (\cref{tab:low-budget-extension}): stronger visual pruning lengthens search, increasing text KV and offsetting visual savings under the native text-retention rules.

\input{tables/18_low_budget_extension}

\subsection{Component ablation}
\label{sec:additional-experiments}
\label{supp:accuracy-provenance}
\Cref{tab:four-component-difficulty} expands the main-paper ablation by difficulty.

\input{tables/21_four_component_difficulty}

Removing source quotas allocates by candidate counts while keeping lineage. Removing lineage uses uniform selection within unchanged quotas. Removing fixed history reselects resident rows at each handoff with the same source/spatial rules, anchors, and original packet tails, without restoring deleted KV. Removing tail protection preserves anchors and the counterfactual capacity floor. All arms retain incremental execution and current-visual selection.

%% file: tables/16_current_budget.tex
\begin{table}[!ht]
\caption{\textbf{Actual total-KV retention for all main-table methods.} Values are percentages of each dataset\textquotesingle s Dense mean.}
\label{tab:current-budget}
\centering\small\setlength{\tabcolsep}{2pt}
\begin{tabularx}{\columnwidth}{l*{5}{>{\raggedleft\arraybackslash}X}}\toprule
Method & VP & V* & 4K & 8K & MME\\
\midrule
Dense & 100.00 & 100.00 & 100.00 & 100.00 & 100.00\\
H$_2$O & 26.03 & 25.72 & 24.77 & 25.21 & 25.33\\
LOOK-M & 24.75 & 24.55 & 24.75 & 25.86 & 30.97\\
VL-Cache & 24.76 & 26.45 & 24.91 & 25.56 & 25.32\\
AirCache & 26.84 & 26.15 & 24.56 & 25.75 & 30.18\\
MixKV & 26.03 & 25.72 & 24.76 & 25.21 & 24.97\\
TGV-KV & 25.62 & 25.83 & 24.74 & 25.31 & 25.32\\
MM-ShiftKV & 25.97 & 25.70 & 25.06 & 25.32 & 25.62\\
AttentionPack & 24.56 & 26.63 & 25.15 & 24.07 & 24.39\\
\midrule
\method{} & 24.34 & 24.72 & 24.47 & 24.21 & 24.63\\
\bottomrule\end{tabularx}
\end{table}

%% file: tables/22_main_difficulty.tex
\begin{table}[!ht]
\centering\small
\caption{\textbf{VisualProbe accuracy by difficulty in the main comparison.} Percentages use all 106 Hard, 268 Medium, and 141 Easy questions. Bold marks the best reported accuracy.}
\label{tab:main-difficulty}
\setlength{\tabcolsep}{4pt}\renewcommand{\arraystretch}{1.12}
\begin{tabularx}{\columnwidth}{l*{3}{>{\centering\arraybackslash}X}}
\toprule
Method & Hard & Medium & Easy\\
\midrule
Dense & 43.40 & 47.76 & 63.83\\ 
H$_2$O & 46.23 & 46.64 & 59.57\\ 
LOOK-M & 34.91 & 44.03 & 59.57\\ 
VL-Cache & 50.94 & 48.13 & 65.96\\ 
AirCache & 40.57 & 44.03 & 54.61\\ 
MixKV & 43.40 & 47.39 & 66.67\\ 
TGV-KV & 33.96 & 42.91 & 63.12\\ 
MM-ShiftKV & 44.34 & 45.15 & 63.12\\ 
AttentionPack & 28.30 & 30.97 & 45.39\\ 
\midrule
\method{} (ours) & \textbf{51.89} & \textbf{50.75} & \textbf{67.38}\\ 
\bottomrule\end{tabularx}
\end{table}

%% file: tables/23_main_rounds.tex
\begin{table}[!ht]
\centering\small
\caption{\textbf{Reasoning rounds in the main comparison.} VP denotes VisualProbe, 4K/8K denote HR-Bench, and MME denotes MME-Lite. Rounds follow the main-comparison speed boundaries defined in Appendix~\ref{supp:runtime-measurement}.}
\label{tab:main-rounds}
\setlength{\tabcolsep}{2pt}\renewcommand{\arraystretch}{1.12}
\begin{tabularx}{\columnwidth}{l*{5}{>{\centering\arraybackslash}X}}
\toprule
Method & VP & V* & 4K & 8K & MME\\
\midrule
Dense & 7.84 & 3.80 & 3.97 & 4.07 & 10.01\\ 
H$_2$O & 8.50 & 18.52 & 5.92 & 4.25 & 9.18\\ 
LOOK-M & 8.40 & 4.80 & 5.35 & 4.36 & 10.00\\ 
VL-Cache & 7.12 & 3.60 & 3.50 & 3.80 & 6.80\\ 
AirCache & 9.19 & 6.57 & 4.42 & 4.80 & 11.25\\ 
MixKV & 6.61 & 3.10 & 3.19 & 3.68 & 5.32\\ 
TGV-KV & 5.56 & 2.71 & 2.99 & 3.37 & 4.23\\ 
MM-ShiftKV & 8.59 & 5.74 & 4.44 & 4.22 & 8.70\\ 
AttentionPack & 10.58 & 4.99 & 4.54 & 4.50 & 7.49\\ 
\midrule
\method{} (ours) & 7.18 & 4.02 & 3.84 & 3.90 & 6.04\\ 
\bottomrule\end{tabularx}
\end{table}

%% file: tables/18_low_budget_extension.tex
\begin{table}[!ht]
\caption{\textbf{Stronger visual pruning can increase realized total KV.} Capacity is the native visual-row budget. All KV percentages use Dense's mean total KV on the same budget-sweep subset as the denominator.}
\label{tab:low-budget-extension}
\centering\small\setlength{\tabcolsep}{3pt}
\begin{tabular}{lrrrr}\toprule
& \multicolumn{2}{c}{AirCache} & \multicolumn{2}{c}{LOOK-M}\\
Capacity (rows) & 128 & 2 & 256 & 2\\
\midrule
Total KV (\%) & 21.10 & 28.41 & 26.20 & 26.85\\
Text KV (\%) & 19.48 & 28.39 & 24.29 & 24.40\\
Visual KV (\%) & 1.61 & 0.02 & 1.91 & 2.45\\
Accuracy (\%) & 28.21 & 5.31 & 10.06 & 3.07\\
Rounds & 12.96 & 19.19 & 17.07 & 18.28\\
Time (s) & 54.20 & 75.80 & 105.98 & 114.15\\
\bottomrule\end{tabular}
\end{table}

%% file: tables/21_four_component_difficulty.tex
\begin{table}[!ht]
\caption{\textbf{Single-component accuracy by difficulty.} Each cell gives correct questions divided by the full difficulty population, followed by accuracy in percent.}
\label{tab:four-component-difficulty}
\centering\small
\setlength{\tabcolsep}{2pt}\renewcommand{\arraystretch}{1.12}
\begin{tabularx}{\columnwidth}{@{}l*{3}{>{\centering\arraybackslash}X}@{}}
\toprule
Configuration & Hard & Medium & Easy\\
\midrule
Dense & \shortstack{46/106\\(43.40)} & \shortstack{128/268\\(47.76)} & \shortstack{90/141\\(63.83)}\\
\method{} (full) & \shortstack{55/106\\(51.89)} & \shortstack{136/268\\(50.75)} & \shortstack{95/141\\(67.38)}\\
w/o source quotas & \shortstack{54/106\\(50.94)} & \shortstack{135/268\\(50.37)} & \shortstack{88/141\\(62.41)}\\
w/o crop lineage & \shortstack{42/106\\(39.62)} & \shortstack{115/268\\(42.91)} & \shortstack{85/141\\(60.28)}\\
w/o fixed history & \shortstack{53/106\\(50.00)} & \shortstack{127/268\\(47.39)} & \shortstack{89/141\\(63.12)}\\
w/o tail protection & \shortstack{44/106\\(41.51)} & \shortstack{115/268\\(42.91)} & \shortstack{85/141\\(60.28)}\\
\bottomrule
\end{tabularx}
\end{table}

%% file: sections/16_cache_growth_appendix.tex
\section{Cache growth during exploration}
\label{supp:cache-growth}
\begin{figure}[!ht]
\centering
\includegraphics[width=\columnwidth]{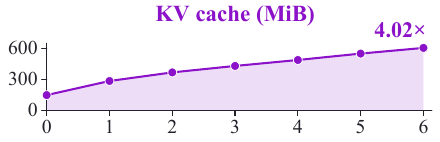}
\par\smallskip
{\small Acquired crops\par}
\medskip
\includegraphics[width=\columnwidth]{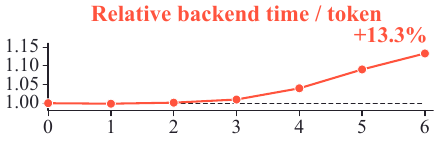}
\par\smallskip
{\small Acquired crops\par}
\caption{\textbf{Exploration increases cache size and per-token cost.} Dense on the 112 VisualProbe questions described in Appendix~\ref{supp:operation-timing-generated}. Top: mean post-prefill KV. Bottom: median per-token backend time, normalized to each question's original-image call. After six crops, KV grows $4.02\times$ and normalized per-token time rises by 13.3\%.}
\label{fig:exploration-growth}
\end{figure}